\documentclass{article}

\usepackage{microtype}
\usepackage{graphicx}
\usepackage{subcaption}
\usepackage{booktabs} %

\usepackage{hyperref}
\usepackage{lineno}

\usepackage[preprint]{icml2026}

\usepackage{amsmath}
\usepackage{amssymb}
\usepackage{mathtools}
\usepackage{amsthm}

\usepackage[capitalize,noabbrev]{cleveref}

\theoremstyle{plain}

\theoremstyle{definition}

\theoremstyle{remark}

\usepackage[textsize=tiny]{todonotes}

\icmltitlerunning{Similarity of Processing Steps in Vision Model Representations}

\begin{document}

\twocolumn[
  \icmltitle{Similarity of Processing Steps in Vision Model Representations}

  \icmlsetsymbol{equal}{*}

  \begin{icmlauthorlist}
    \icmlauthor{Matéo Mahaut}{upf}
    \icmlauthor{Marco Baroni}{upf,icrea}
  \end{icmlauthorlist}

  \icmlaffiliation{upf}{Universitat Pompeu Fabra, Barcelona, Spain}
  \icmlaffiliation{icrea}{ICREA, Barcelona, Spain}

  \icmlcorrespondingauthor{Matéo Mahaut}{mateo.mahaut@gmail.com}

  \icmlkeywords{Representation Learning, Interpretability, Universal Representations, Vision}

  \vskip 0.3in
]

\printAffiliationsAndNotice{}  %

\begin{abstract}
Recent literature suggests that the bigger the model, the more likely it is to converge to similar, ``universal'' representations, despite different training objectives, datasets, or modalities. While this literature shows that there is an area where model representations are similar, we study here how vision models might get to those representations--in particular, do they also converge to the same intermediate steps and operations? We therefore study the processes that lead to convergent representations in different models. First, we quantify distance between different model representations at different stages. We follow the evolution of distances between models throughout processing, identifying the processing steps which are most different between models. We find that while layers at similar positions in different models have the most similar representations, strong differences remain. Classifier models, unlike the others, will discard information about low-level image statistics in their final layers. CNN- and transformer-based models also behave differently, with transformer models applying smoother changes to representations from one layer to the next. These distinctions clarify the level and nature of convergence between model representations, and enables a more qualitative account of the underlying processes in image models.

\end{abstract}

\section{Introduction}

As foundation models grow bigger and more efficient, they solve increasingly general tasks.  These models tend to converge to similar representations as their performance on selected benchmarks increases \cite{pmlr-v44-li15convergent,NEURIPS2018_a7a3d70c, pmlr-v97-kornblith19a, moschella2023relative, pmlr-v235-huh24a}, despite differences in architecture, training data, modality, or training objective. \citet{pmlr-v235-huh24a} argue that this ``universality'' might stem from getting closer to optimal representations of the physical world. 

However, while we now know there is convergence in representation, we still don't know if there is also convergence in the process, that is, the way representations evolve through the different model layers before converging.
As an example, for an easy classification task, which separates inputs according to their shapes and colors, a model could first classify inputs into the different shapes in a first layer, and then extract color information. The opposite strategy would also work: you can get a perfectly separated dataset by separating color first and then separating shapes. Both those processes end up forming the same final representation, but going through different steps. In this example, the intermediate representations for each strategy would be quite distant, even if their final representation is the same.  
We seek to analyze the processing steps of different vision models, to check which processing steps are universally shared, and which are not. We employ quantitative means, looking at the similarity of space at intermediary layers, and then observe how nearest neighbors organize the representational space in vision models.

\subsection{Related work}
\textbf{Representation universals:} Previous work has looked at quantitative differences between different seeds of the same architectures, or different sizes of models \citep{nguyen2020wide, raghu2021vision}, often using CKA as similarity measure \cite{pmlr-v97-kornblith19a}. Specifically, \citet{nguyen2020wide} mentions a "block structure" which might be linked to the "universal reasoning" zones mentioned  in \citet{acevedo2025approachidentifysemanticallyinformative}. They argue those zones with high similarity must be full of redundancies, due to the overparameterization of deep neural networks. In those blocks, there is very little change in performance of different linear probes, and it is possible to remove layers that are part of those blocks with very little impact on model performance. 
Instead, \citet{pmlr-v235-huh24a, acevedo2025approachidentifysemanticallyinformative} observe that these blocks of similar representations preserve semantics across languages and modalities, and \citet{lad2024remarkable} argue that such blocks can be a "feature engineering" phase, where information from middle layers is incrementally added to shape features relevant to later tasks. There is not yet in the literature a complete analysis on the process that leads to the emergence of those zones with similar representations.

\textbf{Shared stages of processing:} It has been established that Convolutional Neural Networks (CNNs) process visual stimuli in a very structured way. We know that CNNs will first find edges and hierarchically combine them into increasingly abstract shapes \citep{zeilerVisualizing,Krizhevsky2017}. This is very similar to the way brains process information \cite{Yamins2016, jocn_a_01544,Suresh2021,  kwon2024braininspired}. 

\textbf{Distance metrics:} Prior work use different metrics to compare datasets and learned representations by characterizing their underlying manifolds, including canonical correlation analysis (CCA) and its variants~\cite{hotelling1936relations,raghu2017svcca}, centered kernel alignment (CKA), which captures nonlinear similarity~\cite{pmlr-v97-kornblith19a}, and information-theoretic or geometry-aware distances measuring statistical dependence or structural differences between feature distributions~\cite{belghazi2018mine,peyre2019computational}.
We choose the \textit{information imbalance} measure, which assumes a local manifold and checks how similar local neighborhoods are in different spaces \cite{pgac039}. Information imbalance is asymmetric (unlike CCA and CKA), non-parametric (unlike CCA), and more stable in high dimensional spaces \cite{acevedo2025approachidentifysemanticallyinformative}.

\subsection{Contributions}
We contribute an extensive quantitative and qualitative cross-layer and cross-model analysis, tracing the evolution of distances between representational spaces, focusing on two examples of the information that might be available at different stages, inspired by the vision literature: low-level features and abstract semantics.
We find that there is structural similarity in the way images are processed, despite some important differences in processing:
(i) \textbf{Shared structure:} Similar model layers occupy the same relative positions, with low-level features consistently preceding abstract semantics; (ii) \textbf{Model differences:} iGPT is more self-similar across layers, ConvNeXt shows erratic layer-to-layer changes, and DINOv2 retains more information in later layers; (iii) \textbf{Feature retention:} Classification models discard low-level features in later layers, whereas self-supervised and next-token prediction models retain them; (iv) \textbf{iGPT behavior:} the next-pixel-prediction model organizes images according to their semantic categories in its middle layers, but loses this structure in its later layers, in line with its more granular objective.

Thanks to neighborhood based methods, we separate what part of processing is universal, from the model-specific characteristics that come from variations in architectures and objectives. This improves understanding of meaning construction in image-based processing.

\section{Methods}

\textbf{Information Imbalance: }To compare representations at different layers of different models, where dimensions might not match, and data might live on a non-linear manifold, we focus on local distances. Information imbalance \cite{pgac039} is a dataset distance metric which looks at how similar two image neighborhoods are. Information imbalance uses the ranked distance of each datapoint with every other one. The idea is to check that, at least locally, the n-nearest neighbors are the same in both of the compared spaces. This is especially relevant in high-dimensional spaces where local geometry is preserved but long-range distances do not behave well, and empirically has been found to work better than CKA in deep networks \cite{cheng2023bridging, acevedo2025approachidentifysemanticallyinformative}. Information imbalance is asymmetric: space A might be able to predict space B, while B could not be used as well to reconstruct space A. This allows us to check if one of the spaces holds more information. Formally, information imbalance is computed as follows: 
{\setlength{\abovedisplayskip}{4pt}
 \setlength{\belowdisplayskip}{4pt}
\begin{equation}
\Delta(A \rightarrow B) \approx \frac{2}{N}\langle r^B \mid r^A = 1 \rangle
\label{eq:ii}
\end{equation}}
With A and B being two spaces, and $r$ being the rank of a point in a given space. $N$ is the number of datapoints considered. For each point we look at the rank in space B of the nearest neighbor in space A. If all neighbors are shuffled, then the information imbalance goes to 1, whereas if order is exactly preserved for all datapoints, information imbalance goes to 0. In practice, lower information imbalance means spaces are more similar.

\begin{table*}[tb]
\centering
\caption{Overview of Vision Models analyzed in this study. Chosen models cover different architecture, parameter count, and training objective. These representative models, with large parameter counts, are used in the main text, while others architectures, sizes and objectives appear in Appendix \ref{app:more-ii}, Table \ref{tab:model_summary_app}.}
\label{tab:model_summary}
\renewcommand{\arraystretch}{1.05}
\begin{tabular}{l l l l}
\toprule
\textbf{Model} & \textbf{Parameter Count (in millions)} & \textbf{Architecture} & \textbf{Training Objective} \\
\midrule
\textbf{iGPT} & 1362.1 & Vision Transformer & Next-pixel prediction \\
\textbf{DINOv2} & 1136.4 & Vision Transformer & Self-supervised distillation \\
\textbf{ConvNeXt} & 350.2 & Convolutional N.N. & Classification (ImageNet) \\
\textbf{ViT} & 304.3 & Vision Transformer & Classification (ImageNet) \\
\bottomrule
\end{tabular}
\end{table*}

\textbf{Models: }We compare vision models covering different architectures, sizes, and training objectives, as summarized in Table \ref{tab:model_summary}. Additional results with other layers, sizes, and architectures are in Appendix \ref{app:more-ii}. ViT \citep{dosovitskiy2020image} and ConvNeXt \citep{liu2022convnet} are different architectures directly trained on supervised image classification. DINOv2 \citep{oquabdinov2} is trained with a mix of self supervision and model distillation. Finally, iGPT \citep{pmlr-v119-chen20s} is trained with a next pixel prediction objective. 

\textbf{Datasets: }Experiments are run using images from the validation set of ImageNet-21K \citep{Ridnik:etal:2021}. We use images from the validation set, which were not seen during training by the classifier models. Additionally, for semantic classification we use the ManyNames Dataset \citep{silberer-etal-2020-humans}, which comes with multiple labels assigned by human annotators to help capture the semantics of an image.

\section{Experiments}
\subsection{Cross-layer model comparisons} 
\label{sec:quant}
In Fig.~\ref{fig:big_ii}, We plot layer-wise model comparisons within and between the largest models of each category. We always report information imbalance comparisons with respect to an  early, a middle and a late layer, having qualitatively ascertained that these are representative of the overall patterns. Information imbalance is estimated using $10^4$ images, which we empirically find to be sufficient for convergence in Appendix~\ref{app:std-ii}. 

\subsubsection{Within-model comparison}
\begin{figure*}[tb]
    \centering
    \includegraphics[width=\textwidth]{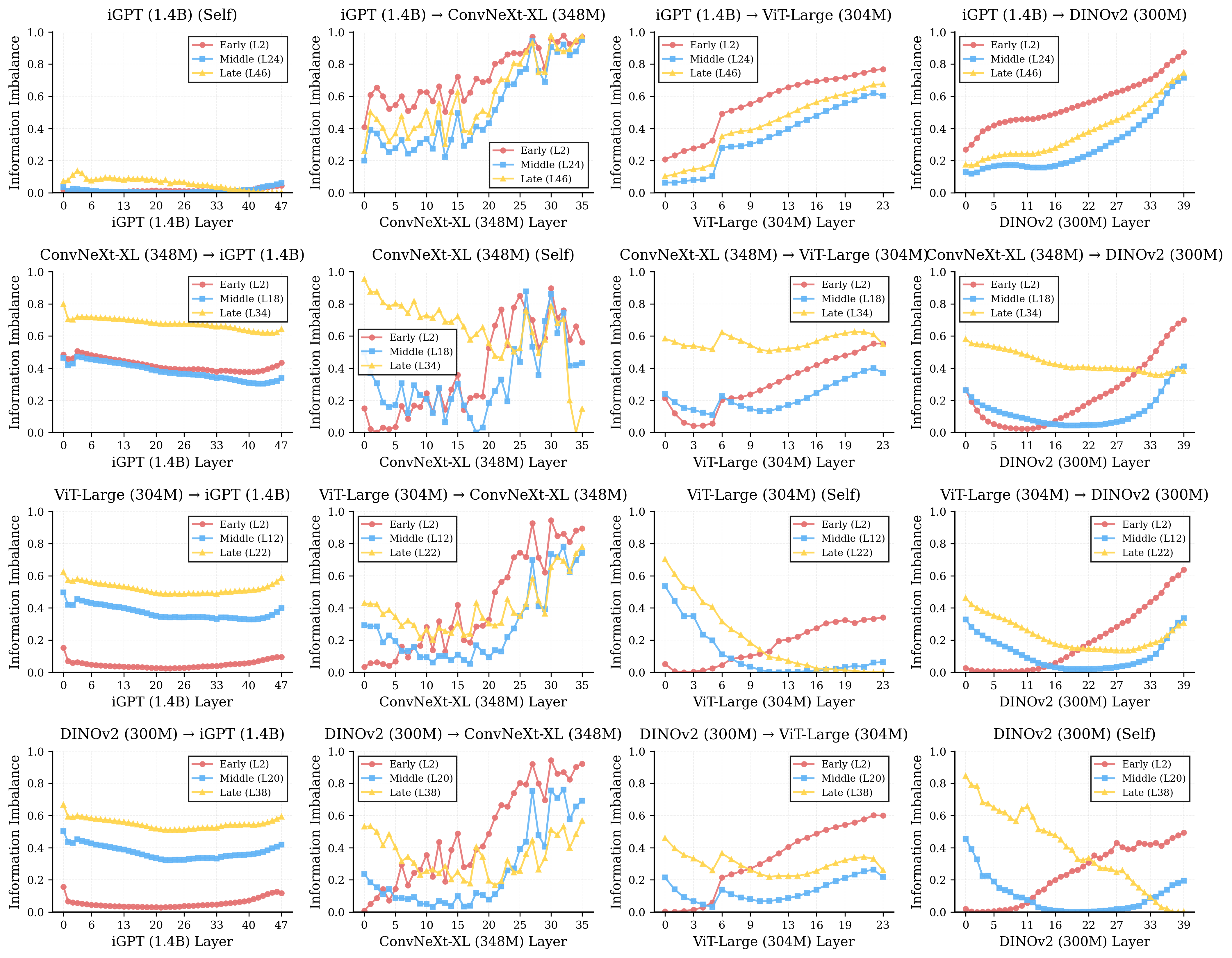}
    \caption{Information imbalance between models at different depths; lower information imbalance for A $\rightarrow{}$ B means the neighbor structure of A is predictive of that of B. Four models are compared, in both directions. We look at the information imbalance for an early (blue circle), middle (purple square) and late layer (yellow triangle) of a first model with all layers of the second model. The title of each subgraph indicates the direction, with an arrow from the predicting model towards the predicted one.}
    \label{fig:big_ii}
\end{figure*}

First, we find that the way images are organized in the latent space of models \textbf{changes incrementally throughout the layers}.  The subplots on the \textit{diagonal }of Fig.~\ref{fig:big_ii} show information imbalance comparisons of different layers within the \textit{same} models. We find that information imbalance mostly incrementally increases as distance grows between compared layers. %
DINOv2 (bottom right) is the clearest example: its information imbalance with respect to early and late layers is linearly correlated to between-layer distance. Information imbalance of the middle layer is not directly linear to distance, but cues instead a wider area of strong layer-to-layer similarity. %
ViT follows a similar process, with a floor effect where middle layers stay similar to later layers. 

\textbf{Information imbalance broadly follows the same profile but is not smooth for ConvNeXt} (second row, second column of Fig.~\ref{fig:big_ii}). Unlike for DINOv2 or ViT, %
which have smooth variation in information imbalance across layers, 
the transition in information imbalance from one layer to the next is irregular for ConvNeXt. Taking the middle layer as an example (blue squares), on layer 5 information imbalance is around 0.2, then it jumps to 0.3 at layer 6, goes back below 0.2 at the next layer, then it raises again, etc. In Appendix \ref{app:std-ii}, we verify that this is not an effect of sample size, as well as extend the information imbalance analysis to multiple samples of the same dataset, with similar results. The standard deviation of adjacent-layer information imbalance differences is always at 0.1 or higher, compared to $<0.05$ for the other models. We propose an explanation for this in Section \ref{sec:convn}.

\textbf{All iGPT layers have very low information imbalance} ($\leq 0.15$). The model has a very wide zone of very similar layers. However, even at this scale the most distant layers are the least similar, as was the case with the other models. The late layer is least similar to the early layers, and the middle layer forms an (albeit rather flat) U shape.

Overall, we find a generally intuitive pattern. Whenever an image representation goes through a layer, the representation is modified. The more layers it goes through, the more different it becomes. We also observe zones where representations remain very similar for many layers in the middle layers. Finally, each model has its own specificities, with information imbalance not being smooth for ConvNeXt, ViT displaying a flooring effect, and iGPT having very low information imbalance for all layers.

\subsubsection{Cross-model comparison}
\label{sec:between-models}
\begin{figure*}[tb]
    \centering
    \includegraphics[width=1\linewidth]{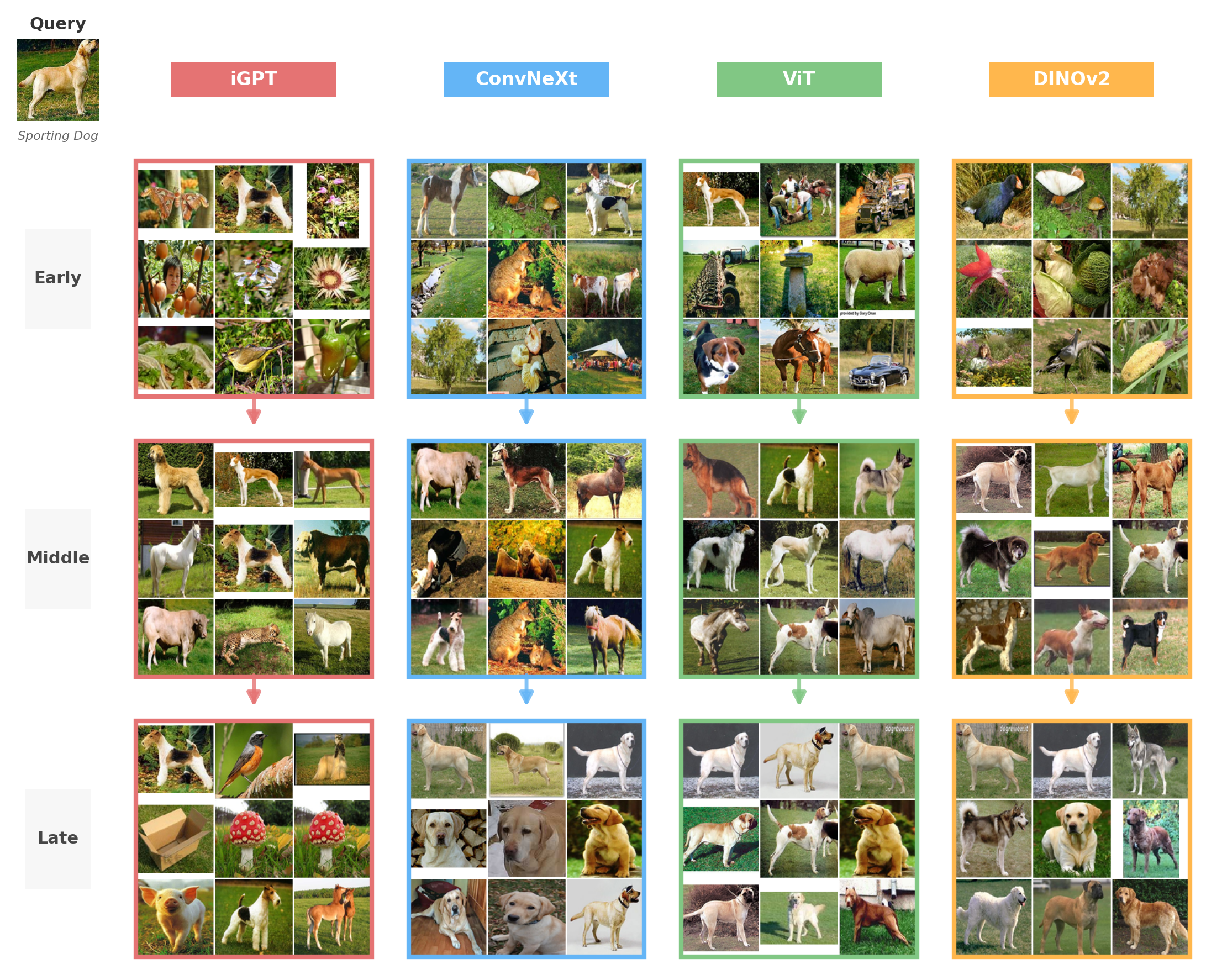}
    \caption{Nearest neighbors of a \textit{sporting dog} (top left) image from a set of 100k images, computed using cosine similarity in the space of different layers for different models. Each column is a model, each row is a specific layer. Early layers are all second layers, and late layers are all penultimate layers to avoid effects from tokenization or detokenization on the very first and very last layers, respectively.}
    \label{fig:nn_dogs}
\end{figure*}
Relative layer distance remains an important similarity predictor even between models. Early layers from one model are most similar to early layers of other models, middle to middle, and late to late. We find areas of convergence for a majority of model pairs, confirming what was found in recent literature. However, we also observe different patterns in similarity profiles, suggesting distinct processes. %

We first look at X $\rightarrow$ DINOv2, that is, layers of other models predicting those of DINOv2, found in the last column of Fig.~\ref{fig:big_ii}. We compare them with DINOv2 $\rightarrow$ X, DINOv2 predicting other models, in the last row of Fig.~\ref{fig:big_ii}. DINOv2' late layers (after 33) are the hardest to predict, with a notable bump in information imbalance after that point for all models. Those final layers consistently have information imbalance values above 0.4, and even reach 0.92 (early iGPT predicting DINOv2's layer 35). %
On the other hand, the late layer of DINOv2 (yellow triangles) predicts iGPT image ranks with an information imbalance of 0.6 or lower, well below 0.92, and ViT layers with information imbalance of 0.4 or lower, instead of 0.6, as in the other direction. For ConvNeXt, both directions have equivalent information imbalance, around 0.5. This suggests that, while DINOv2 holds information that is predictive of different layer spaces in other models, the latter lack information to do the same. \textbf{DINOv2 holds more information than other models in its late layers}.

Another notable asymmetry is with the ConvNeXt model. Predicting the ConvNext model (second column of Fig.~\ref{fig:big_ii}) results in information imbalance curves which are not smooth. Nonetheless, \textbf{ConvNeXt layers predict other models smoothly} (second row of Fig.~\ref{fig:big_ii}). If, as previously seen, adjacent layers of a ConvNeXt model do not stably predict each other, then another model predicting one of its layers cannot reliably predict the next one either. A single layer of ConvNeXt can therefore predict incremental variations in other models smoothly, but \textbf{another model will face the same irregularities previously mentioned when predicting ConvNeXt}.

While each model has its own peculiarities, the ``distance rule'', whereby layers similarly situated in different models predict each other best, generally holds true. The iGPT model, however, is a counter-example. It displays fundamental differences in the way it processes images. All its layers share the same, or a very similar structure. This results in all iGPT layers being equally predictable by another model's layer, independently of their position. In the first column of Fig.~\ref{fig:big_ii}, this results in near constant information imbalance depicted by almost horizontal lines. \textbf{A given model layer can predict all iGPT layers equally well}. The common structure to all\textbf{ iGPT layers is most similar to the one in early layers of other models}. The early (red) ViT and DINOv2 layers both have the lowest information imbalance to predict any iGPT layer, while for ConvNeXt both the middle and early layers predict iGPT similarly well.  Specifically, iGPT's middle layers are particularly close to the first 5 layers of the ViT architecture.

\subsection{Exploring image neighborhood organization}
\label{sec:organisation}

 In Sec.~\ref{sec:between-models}, we identified sections where models show strong differences. We look next at what happens to image neighborhoods in those regions of interest. Looking at image neighborhoods, we consistently replicate the early CNN-based findings \cite{zeilerVisualizing,Krizhevsky2017} that \textbf{models have a progressively more semantic organization of images.} 
 Fig.~\ref{fig:nn_dogs} shows the 9 nearest neighbors of an image of a dog in Imagenet-21k's \textit{Sporting Dog} category, in the space of an early, middle, and late layer of each model. Qualitatively we notice that \textbf{in the early layers} some low-level features are shared across images. The ViT model already includes a few other four-legged mammals in its neighborhood, and all neighbors for all models share a green natural background, often a lawn. \textbf{The middle layer} neighborhoods are already  more semantically organized. All models show neighbours that are four-legged mammals, often dogs. It should be noted that, still, these animals are often in the same position as the original dog in the query image. \textbf{In the final layers}, neighborhoods only group dogs, often of the same breed as the query image. The position of the dog is not necessarily the same, and the background color is not always a lawn, especially for ConvNeXt. The organisation of the space is increasingly abstract, and centered on meaning. \textbf{The notable exception is iGPT, which regresses to a neighborhood with no clear semantic similarity}. iGPT is also the only model that keeps one specific dog in all three neighborhoods. The second nearest image in the early layer is the same as the fifth in the middle layer, and first in the late layer. This is coherent with the results above showing that iGPT has the lowest inter-layer information imbalance. Additionally, two out of three of the top neighbors (first row of a given neighborhood) are shared for all late layers, except iGPT, showing all but this model to have converging representation spaces.
 
Three additional representative neighborhood examples are in Appendix \ref{app:neighbors}. They confirm that early layers have low-level similarities which become increasingly semantic as we move to later layers. The iGPT model holds less semantics in its later layers than other models, and shares most neighbors across layers, while other models seem partly converge in the final steps. Additional out-of-domain probing experiments can be found in App.~\ref{app:probing}.

\subsubsection{Low-level features}
\begin{figure}[tb]
    \centering
    \includegraphics[width=1\linewidth]{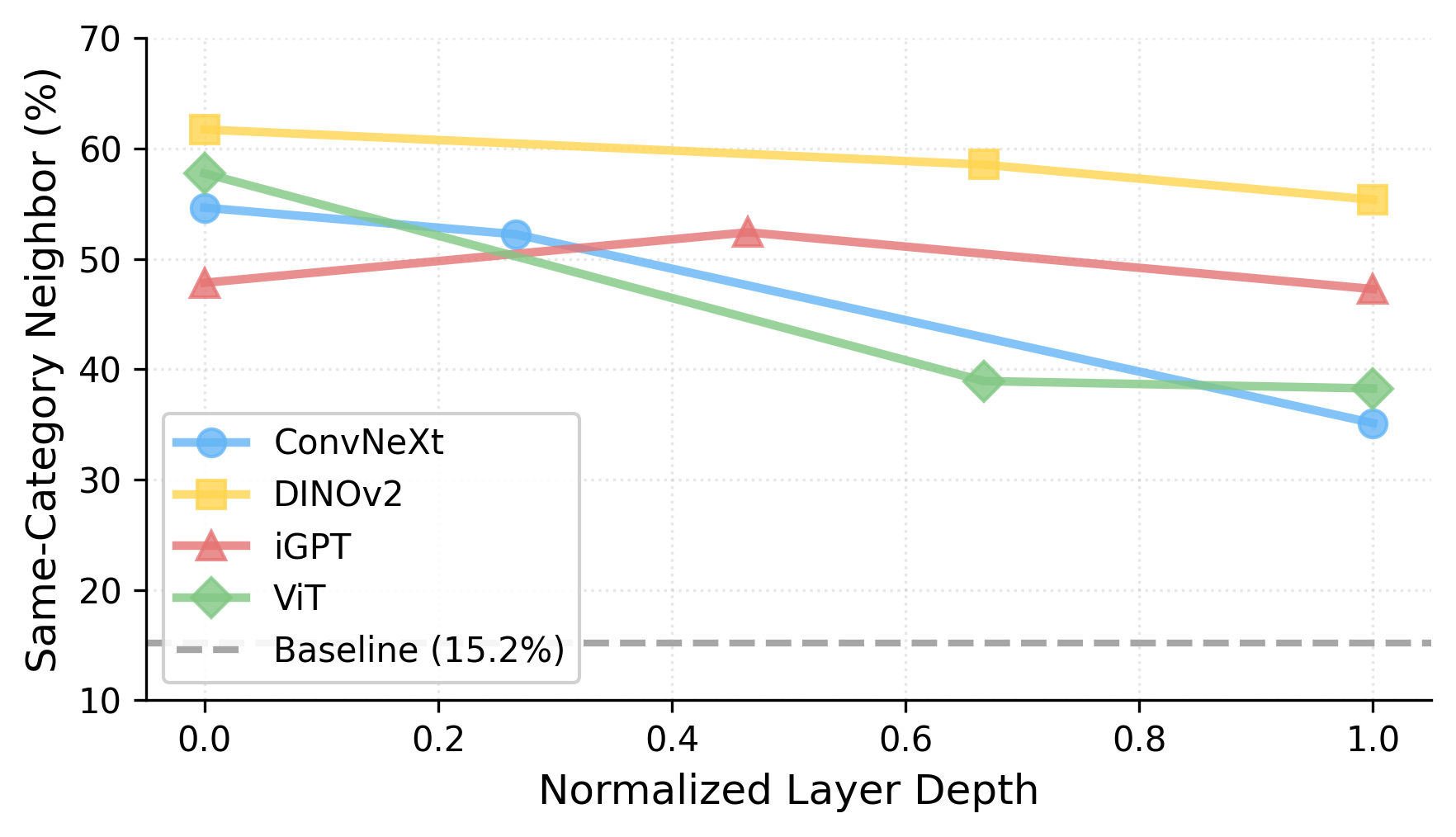}
    \caption{Percentage 10 nearest neighbors of an image that share with it at least one of 9 categories of low-level features. Baseline is for the case where images are randomly spread in the space.}
    \label{fig:low-level}
\end{figure}

To test whether early layers organize representations around low-level features, we labeled a set of images for 3 such features. %
We go through the dataset and compute for each image these 3 properties:
1) \textbf{Edge density}, measured using a Canny edge detector that computes the proportion of pixels which are edges. 2) \textbf{Color warmth}, computed through a comparison of red and blue channel intensity, averaged across the image. 3) \textbf{Texture complexity}, using a Sobel filter to compute the standard deviation intensity changes between pixels. Low variance corresponds to smooth images, while images with rough textures will have high variance. We use 2k images from the Imagenet-21k validation set, and discretize the three properties into three features each, by clustering together the 100 top, 100 middle, and 100 minimum values. Each of these 100 image groups will therefore have similar values for one low-level feature. Some images might belong to multiple categories (e.g., a sunset by the sea might have both warm colors and smooth textures). We estimate that, if images were completely randomly distributed in the space, then they should have the same class  as 15.2\% of their neighbors on average (slightly above the 11.1\% we would get if there was no overlap).

In Fig.~\ref{fig:low-level}, we use the 900 images obtained in this way  to probe where models rely on the relevant low-level information when processing an image. We show how many of an image's nearest neighbors share at least one of the 9 low-level categories. On par with previous works on convolutional architectures \citep{zeilerVisualizing,Krizhevsky2017, jocn_a_01544,kwon2024braininspired}, we find that early layers are largely organized around low-level features, with more than 48\% neighbors of each image being from the same category for all models. This quantity decreases significantly in later layers of models trained as supervised classifiers (ViT and ConvNeXt). Both DINOv2 and iGPT keep instead strong low-level feature sensitivity until the late layers. iGPT is the only model with an \textit{increase}, reaching 51\% neighbors with matching categories in its middle layer, before a slight decline in the later layers. In App.~\ref{app:low}, we repeat the analysis on each property independently, finding that, for all model, color is very important to neighborhood organization, but especially so for iGPT. 

Relating this to the dog example in Fig.~\ref{fig:nn_dogs}, for the late layers we do note that iGPT and DINOv2 neighbors only depict outdoor backgrounds, mostly with green grass of similar texture, while ViT has more background color variation, and ConvNeXt even more, with logs and wooden floorboards, showing how those models categorize less based on low-level features in their late layers, where instead we notice coherence at the semantic level.

\subsubsection{Semantic Coherence}
We look at neighborhoods of images from the ManyNames dataset \cite{silberer-etal-2020-humans}, providing multiple annotations for each image, as elicited from different annotators. This allows us to verify semantic coherence as it aligns to a set of human annotators. We randomly select 50 images and their 10 nearest neighbors for early, middle and late layers in each model, and verify how aligned the labels of those neighborhoods are. We use Jaccard Index, where we divide the amount of shared labels by the total amount of labels for a given pair, to get an idea of how similar the label sets of each of the 50 image neighborhoods are. When images share a lot of labels, it means that models group together the same images that are clustered by human annotations, presumably showing semantically coherent neighborhoods.

\begin{figure}[tb]
    \centering
    \includegraphics[width=1\linewidth]{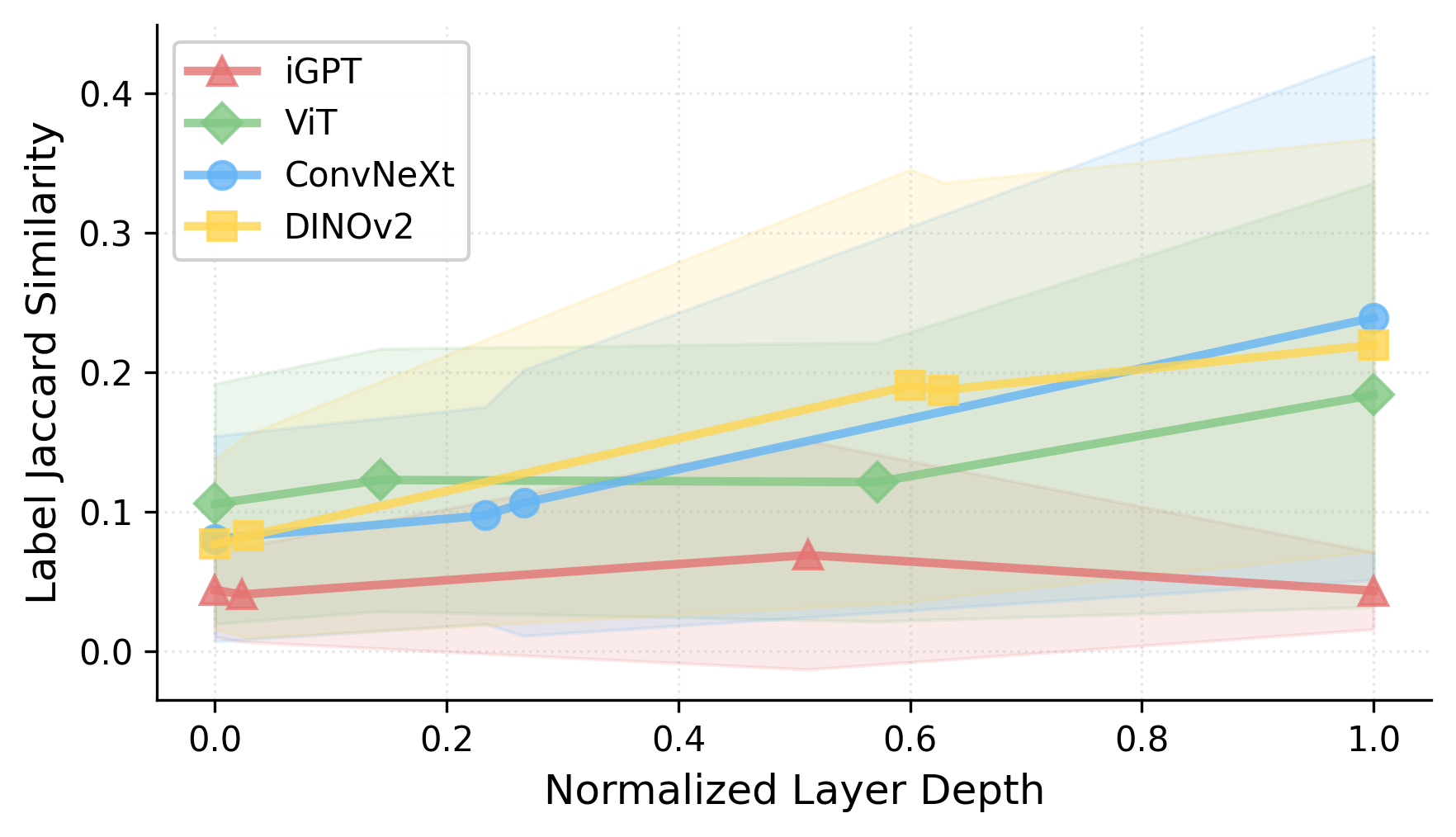}
    \caption{Shared ManyNames labels divided by total labels (Jaccard Similarity) for 50 randomly sampled images and their neighborhoods. Higher Jaccard similarity means that a given neighborhood shares more labels. Shaded out regions show standard deviations across the 50 images and their 10 nearest neighbors.}
    \label{fig:semantic}
\end{figure}

Fig.~\ref{fig:semantic} shows the opposite trend of Fig.~\ref{fig:low-level}. As we progress through layers, images are organized in increasingly semantically coherent neighborhoods. This is true for all models except iGPT, which has neighborhoods with little semantic coherence across the layer span, with again a maximum in its middle layers. DINOv2, on the other hand, keeps both a high amount of low-level information \textit{and} of semantic information until its final layers. Once again, when relating this to Fig.~\ref{fig:nn_dogs}, we note that early-layer neighborhoods are semantically incoherent, and groupings improve as we get to the later layers, with the exception of iGPT, for which middle layers show the best semantic coherence.

\subsubsection{ConvNeXt layer-to-layer similarity}
\label{sec:convn}
In Fig.~\ref{fig:big_ii} we found wide differences how well adjacent ConvNeXt layers could predict each other. This is specific to that model, given that the differences are instead increasing smoothly with layer distance for the others. We try to assess if these information-imbalance irregularities are due to strong cross-layer differences in information content.
In particular, we verify if the semantic information available to a probe at a given layer is missing in the next. We train layer-by-layer linear probes to detect a specific class from the ImageNet dataset \citep{ILSVRC15}. Each probe is a class-specific binary classifier. Training is done on the validation set, i.e., on images which were not seen at training time by the models trained with a classification objective. Results are computed on 10\% held-out images from the same set.

\begin{figure*}[tb]
    \centering
    \includegraphics[width=0.75\linewidth]{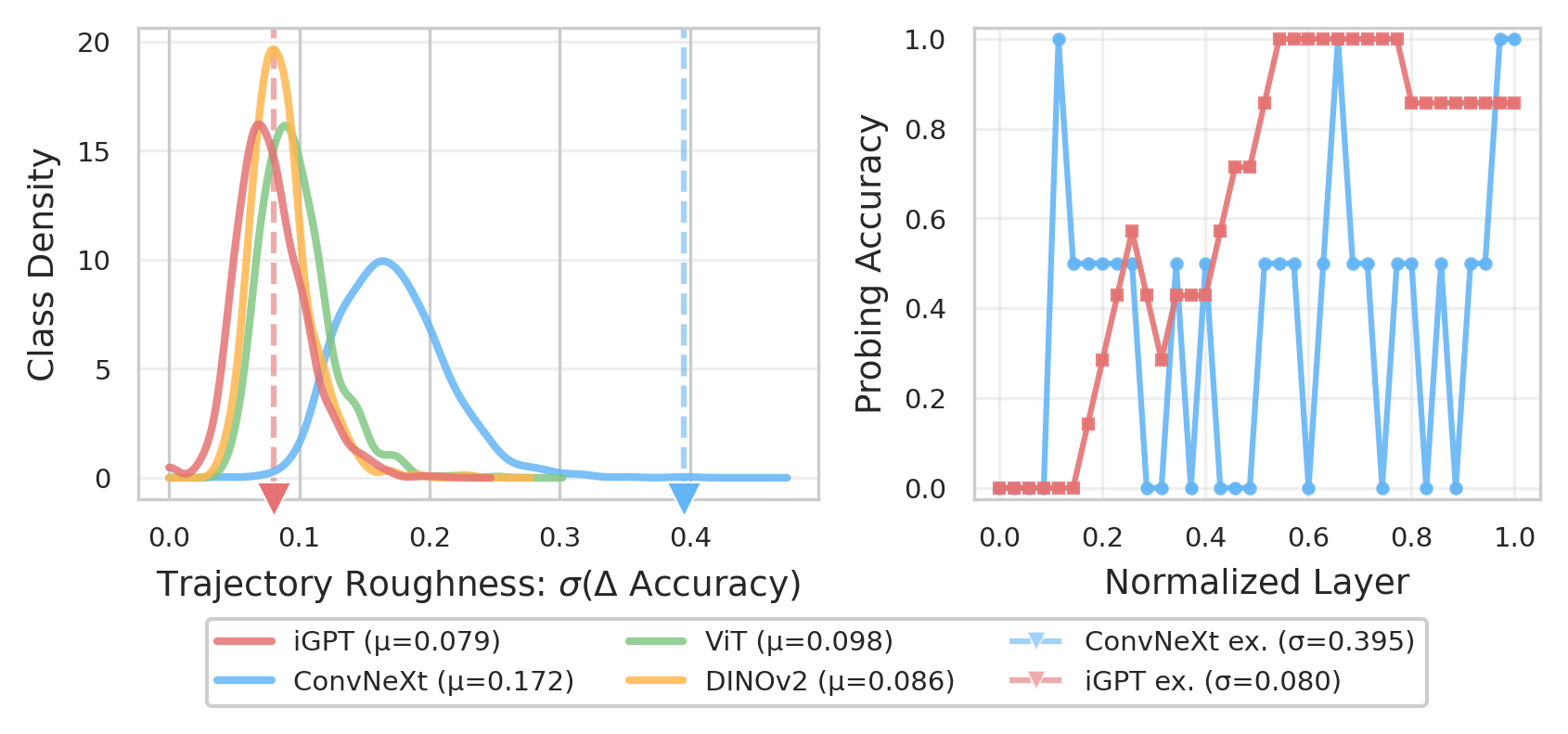}
    \caption{Density plot of classification accuracy roughness across layers (\textit{left}), and examples (right) of smooth (red iGPT example) and rough (blue ConvNeXt example) probing accuracies across layers. We consider a curve ``rough'' if its cross-layer changes are inconsistent, which we measure with standard deviation of layer-to layer differences. In contrast, a smooth curve would consistently have the same layer-to-layer variation. The two example classes have trajectory roughness of 0.395 (\textit{pizza}) and 0.080 (\textit{bloodhound}), respectively, which are marked in dashed lines on the density plot.}
    \label{fig:roughness}
\end{figure*}

Fig.~\ref{fig:roughness} shows two example classes on the right, one in red from the iGPT model, and one in blue, which is an extreme example from the ConvNeXt model. Following the red curve from left to right, we see that, as we use probes trained on representations from deeper inside the iGPT model, accuracy increases. There is a small drop in performance between 0.2 and 0.3 model depth, and another small accuracy decrease around 0.8. More surprisingly, for the blue curve, accuracy is much more chaotic. Instead of a a gradual increase, accuracy peaks early on to 100\%, before decreasing to 50\% for a few layers, and regularly spiking back up to 50\% or 100\%. In this example, a linear probe being able to correctly detect class information on a layer does not inform us about wether its neighboring layers will have similar performances. 

In general, accuracy trajectories with high standard deviation in layer-to-layer accuracy differences are \textit{rough}, as in the blue example, and low standard-deviation ones are \textit{smooth}, as with the red example. On the left of Fig.~\ref{fig:roughness}, we look at the distribution of trajectory roughness estimated in this way for every class in every model. iGPT and ViT have the most classifier smoothness. ConvNeXt has a much wider distribution, with more classes following rough trajectories throughout model layers. To complete the analysis, we verified that, for this model, different classes have their highest probing accuracy on different layers, resulting in the latter being strongly specialized in different subjects.%

We thus confirmed what first observed for ConvNeXt through information imbalance profiles. Instead of incremental modifications of the representation space following changes in organization of semantics, its variations are more radical, going back and forth between specialized layers.  Convolutional neural networks such as ConvNeXt typically display changes linked to the discrete composition of their locality bias, as well as compression of their convolution filters \cite{raghu2021vision,sharon2024representationdynamics}. On the other hand, models with architectures based on attention tend to exhibit a smoother and more uniform evolution of representation geometry across layers than convolutional neural networks. ViT layers remain highly correlated across depth, indicating gradual and coherent geometric transformations of the feature space \cite{raghu2021vision,jiang2024representationprogression}. This difference is attributed to self-attention’s global relational updates, which propagate context uniformly across layers, versus convolution’s locally compositional and hierarchical feature construction \cite{madureira2023vitcnnreview}.

\section{Discussion}

In line with previous work, we find a zone of strong representation similarity in all models, characterized by low information imbalance in Fig.~\ref{fig:big_ii}. There is furthermore an effect of relative depth, with layers at similar depths being most similar. In Sec.~\ref{sec:organisation}, we confirmed that, just as originally observed for supervised CNNs, early layers capture low-level features and later layers more semantic ones, extending the result to other architectures and training regimes. %

We showed, however, that this process is not universal,  with iGPT constituting a notable exception. In the case of iGPT, images remain organized according to low-level statistics, while semantic organization never becomes aligned with human annotations. DINOv2 is also different, as it preserves \textit{both} low-level image statistics and abstract semantic content, while for classification-trained models, low-level similarity decreases with layers. This can be linked to the asymmetry in information imbalance observed in Sec.~\ref{sec:between-models}. The extra information found in the late layers of DINOv2, which ViT and ConvNeXt struggle to predict, might at least in part be due to low-level features, such as edge or color properties, that DINOv2 preserves until the very end.

The patterns might be \textbf{explained by the differences in training objective}. For iGPT, the final task of next pixel prediction is less affected by semantic content and more by nearby pixels in the image. Semantic information is nonetheless useful at the image level, and is still useful for pixel selection. Correspondingly, semantic information is only lost in late layers, while low-level information is also preserved there. The middle layers, where semantic information is more clearly present, are where iGPT best predicts other models. Models trained with classification objectives, such as ConvNeXt and ViT, do not need to carry low-level feature information once abstract semantic information has been acquired, as this is sufficient for classification. Finally, for DINOv2, the mix of self supervision and distillation calls for keeping both types of features. The discriminative part of the loss will lead the model to retain more granular details, beyond class information. This is also the largest model, with the highest dimension for hidden layers, allowing for more information to be kept.

\textbf{Architecture-driven inductive biases also affect processing}. We find that ConvNeXt has stronger layer-to-layer differences in information content. This difference also corresponds to variations in available semantic content, as shown in Fig.~\ref{fig:roughness}. This leads to sequential layers being better or worse at predicting categories. Variation of the image neighborhoods is not as smooth as with attention-based architectures. Note that, in convolutional architectures, all features are local combinations of previous layers. This locality constraint does not exist in attention heads, which can more easily attend the entire pixel space. This forced locality in convolutional networks might be the cause of the observed layer-by-layer variations.

As studies increasingly focus on universality, it is important to clarify which kind of convergence is happening (semantic, low-level, etc.) as well as to distinguish representational and processing convergence. This will allow a better and more qualitative account of the processes behind image understanding, hopefully leading to more controllable and efficient models, and shedding light on the universal and system-specific properties of this cognitive ability.

\bibliography{main}
\bibliographystyle{icml2026}

\newpage
\onecolumn
\subsubsection*{Acknowledgments}
Our work was funded by the European Research Council (ERC) under the European Union’s Horizon 2020 research and innovation programme (grant agreement No. 101019291). This paper reflects the authors’ view only, and the ERC is not responsible for any use that may be made of the information it contains. The authors also received financial support from the Catalan government (AGAUR grant SGR 2021 00470). 
We thank the members of the COLT team at UPF for their feedback throughout the project. We thank Alessandro Laio and Santiago Acevedo for their help understanding early information imbalance results. 
\section*{Impact statement}

This work examines whether the convergence observed in representations of vision models at specific depths is also reflected in the \textbf{intermediate processing steps} that lead to those representations. By analyzing similarities and differences in \textbf{intermediate representational spaces} across models, our results contribute to a more detailed understanding of how high-performing vision systems process visual information.

A better characterization of \textbf{shared and divergent processing steps} may support improved interpretability and more informed model comparison, and may guide future research on architecture design, transfer learning, and evaluation practices that consider internal representations in addition to task performance. Identifying common and divergent processing patterns could also reveal shared vulnerabilities and caution against overly strong assumptions about model equivalence in settings where important differences remain.

This work is \textbf{primarily empirical and descriptive}, and does not establish that any observed convergence in processing is optimal or necessary. Further research is needed to understand how these findings relate to robustness, fairness, and safety in real-world applications.

\appendix
\clearpage

\section{More information imbalance, for more models, at more layers}
\label{app:more-ii}
In Table \ref{tab:model_summary_app}, we list additional models on which we repeat our information imbalance analysis. 
\subsection{DINO \& ViT}
\begin{figure}
    \centering
    \includegraphics[width=1\linewidth]{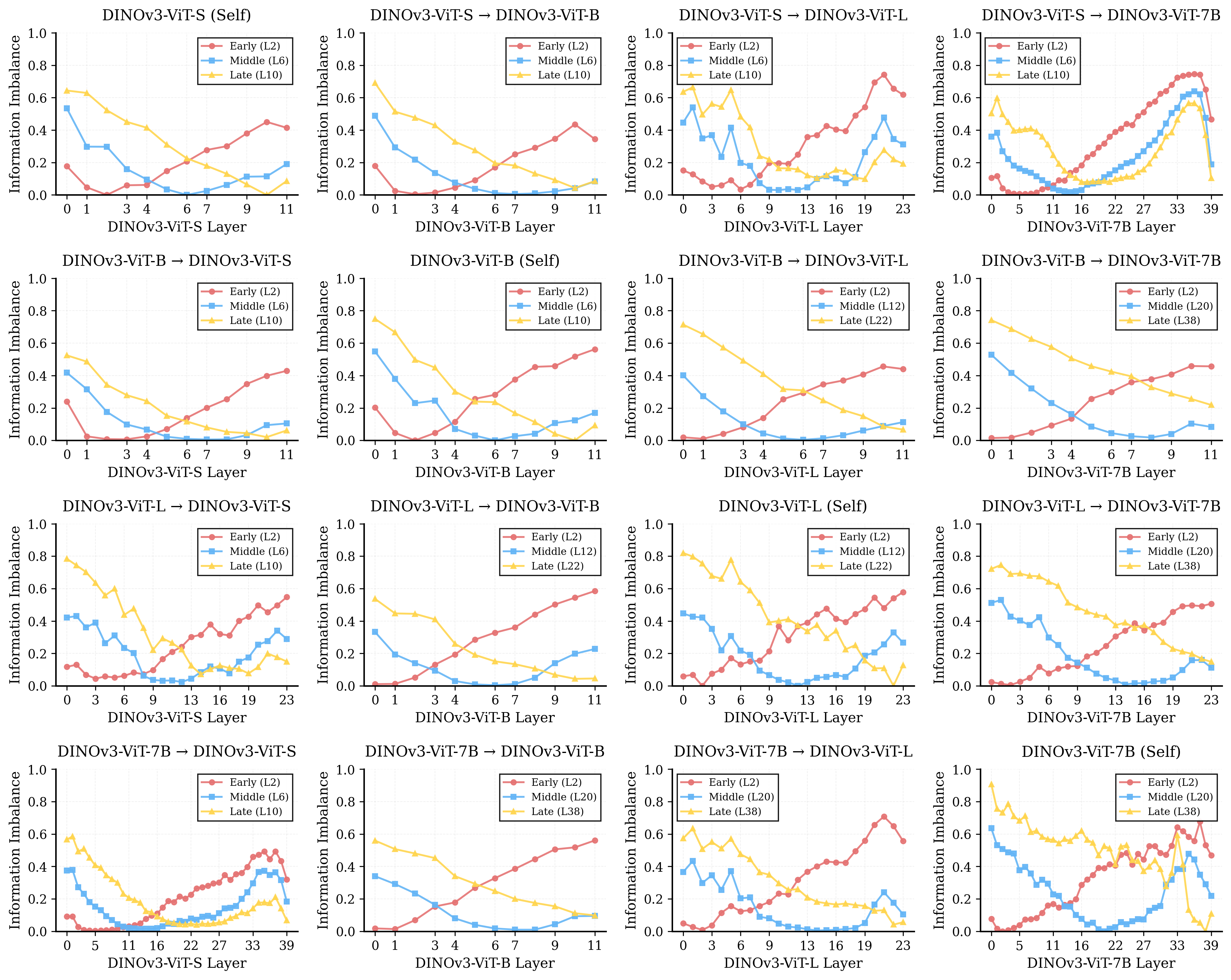}
    \caption{Information imbalance between models for selected representative layers of \textbf{DINOv3 models of different sizes}; lower information imbalance means that more neighbors are shared across image representations on the corresponding layers. Four models are compared, in both directions.  The title of each subgraph indicates the direction of information imbalance, with an arrow from the predicting model towards the predicted one.}
    \label{fig:dino-size-ii}
\end{figure}

We use DINOv3 \cite{simeoni2025dinov3}, which is trained similarly to DINOv2, but on much more data, and with different architectures and sizes. Slight variations to the training regime, such as Gram Anchoring, or high resolution fine-tuning, were also added. We take advantage of the existence of different model sizes to evaluate the effect of size in Fig.~\ref{fig:dino-size-ii}. We find that size does not strongly impact the trend: all DINOv3 models follow the distance rule, where relatively closer layers are most similar.

\begin{figure}
    \centering
    \includegraphics[width=1\linewidth]{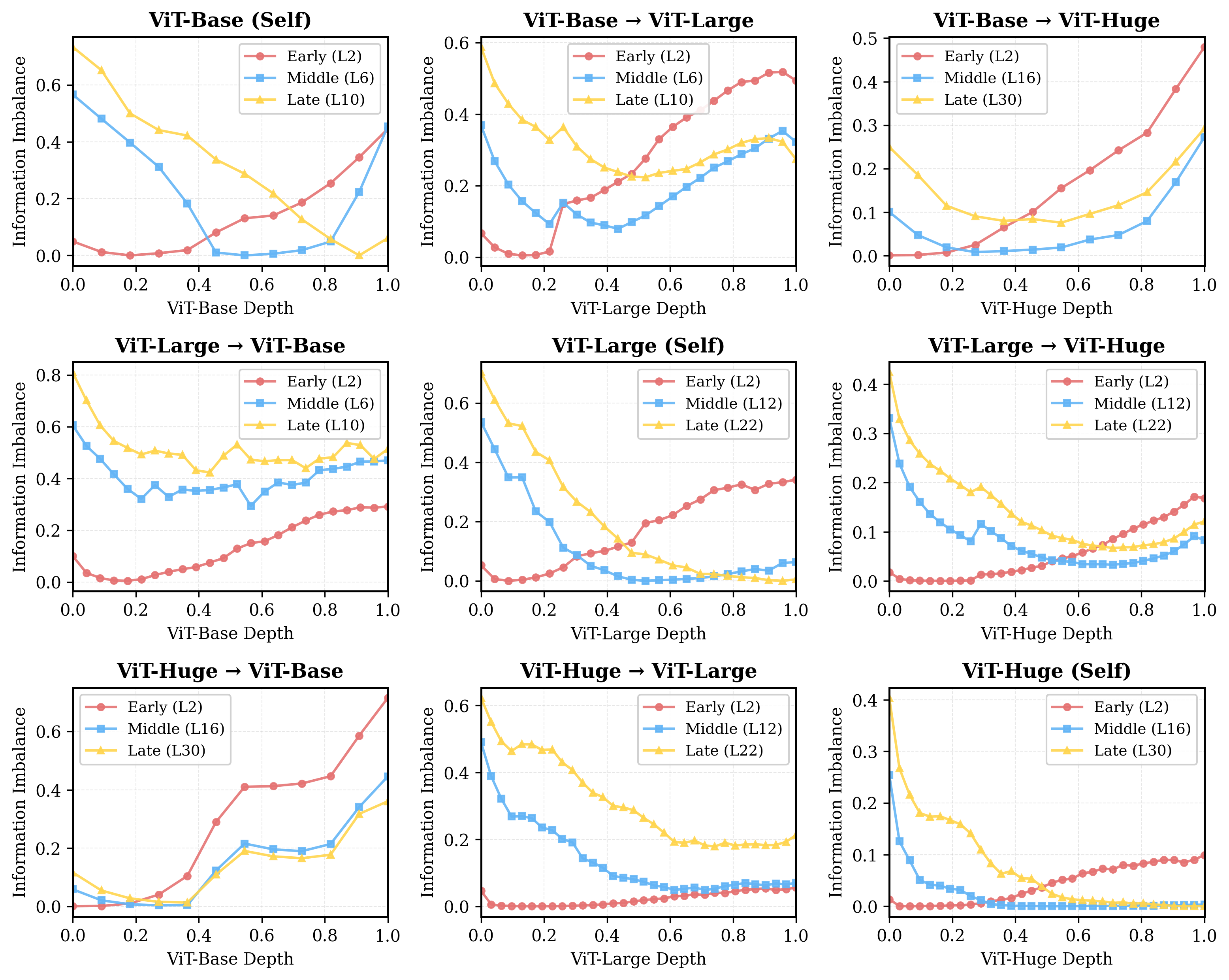}
    \caption{Information imbalance between models for selected representative layers of \textbf{ViT models of different sizes}; lower information imbalance means that more neighbors are shared across image representations on the corresponding layers. Four models are compared, in both directions.  The title of each subgraph indicates the direction of information imbalance, with an arrow from the predicting model towards the predicted one.}
    \label{fig:vit-type-ii}
\end{figure}

The same can be seen for the different sizes of ViT models in Fig.~\ref{fig:vit-type-ii}, with ViT-B middle and later layers being overall harder to predict for the other sizes.

\subsection{ConvNeXt}
\begin{figure}
    \centering
    \includegraphics[width=1\linewidth]{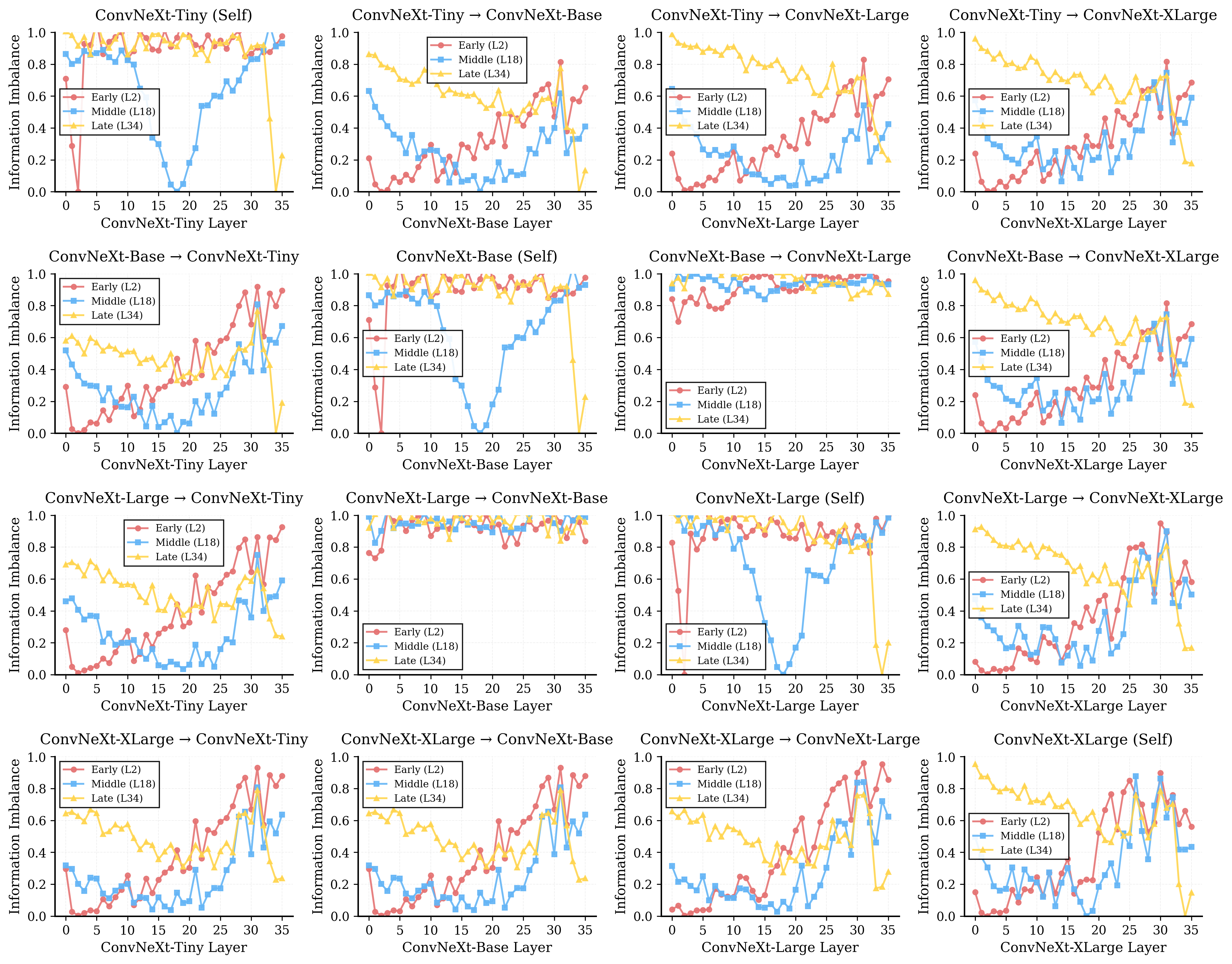}
    \caption{Information imbalance between models for selected representative layers of \textbf{ConvNeXt models of different sizes}; lower information imbalance means that more neighbors are shared across image representations on the corresponding layers. Four models are compared, in both directions.  The title of each subgraph indicates the direction of information imbalance, with an arrow from the predicting model towards the predicted one.}
    \label{fig:conv-size-ii}
\end{figure}

We add multiple ConvNeXt models, with different sizes, and with training variations. DINOv3-ConvNeXt is a ConvNeXt architecture trained with a DINOv3 objective \cite{simeoni2025dinov3}, while ConvNeXt V2 \cite{Woo_2023_CVPR} is similar to the ConvNeXt architecture, but with \textit{Global Response Normalization} to improve feature quality and with an FCMAE training objective, an adaptation of the masked auto-encoder paradigm for convolutional neural networks. In Fig.~\ref{fig:conv-size-ii} we compare the different ConvNeXt sizes. Strong size effects appear with ConvNeXt, unlike with DINOv3. Along the diagonal, the three smallest models do not have incremental variations in neighborhood, but strong drops of up to 0.9. early and late layers are only similar to their one nearest layer. For the middle layer, there is a wider similarity zone, but once again it is radically different from the first ten and last ten layers.

\begin{figure}
    \centering
    \includegraphics[width=1\linewidth]{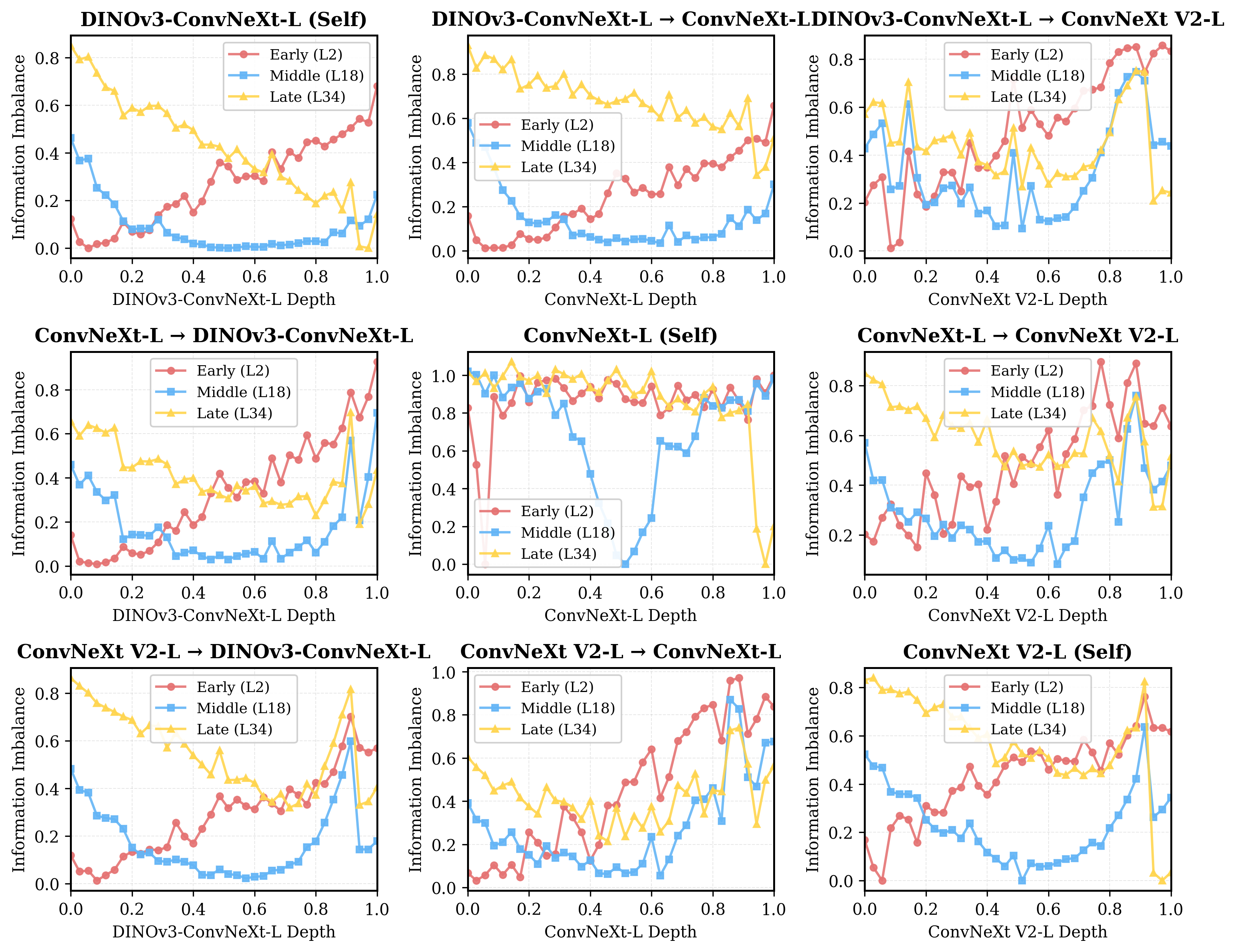}
    \caption{Information imbalance between models for selected representative layers of \textbf{ConvNeXt models of different types}; Dinov3-ConvNeXt-L is a ConvNeXt model trained with the Dinov3 paradigm, ConvNeXt V2 has slight architectural changes. lower information imbalance means that more neighbors are shared across image representations on the corresponding layers. Four models are compared, in both directions.  The title of each subgraph indicates the direction of information imbalance, with an arrow from the predicting model towards the predicted one.}
    \label{fig:conv-type-ii}
\end{figure}

In Fig~\ref{fig:conv-type-ii} we compare the different training regimes. DINOv3-ConvNeXt self-similarity and ConvNeXt V2 self similarity are markedly smoother than ConvNext-L, or XL (see Fig.~\ref{fig:big_ii}), but not as much as the DINOv3, DINOv2, or ViT models. Further testing is required to identify whether the amount of data or the different regularisation in the training loss contributed to increased smoothness.

\subsection{More layers}
\begin{figure}
    \centering
    \includegraphics[width=1\linewidth]{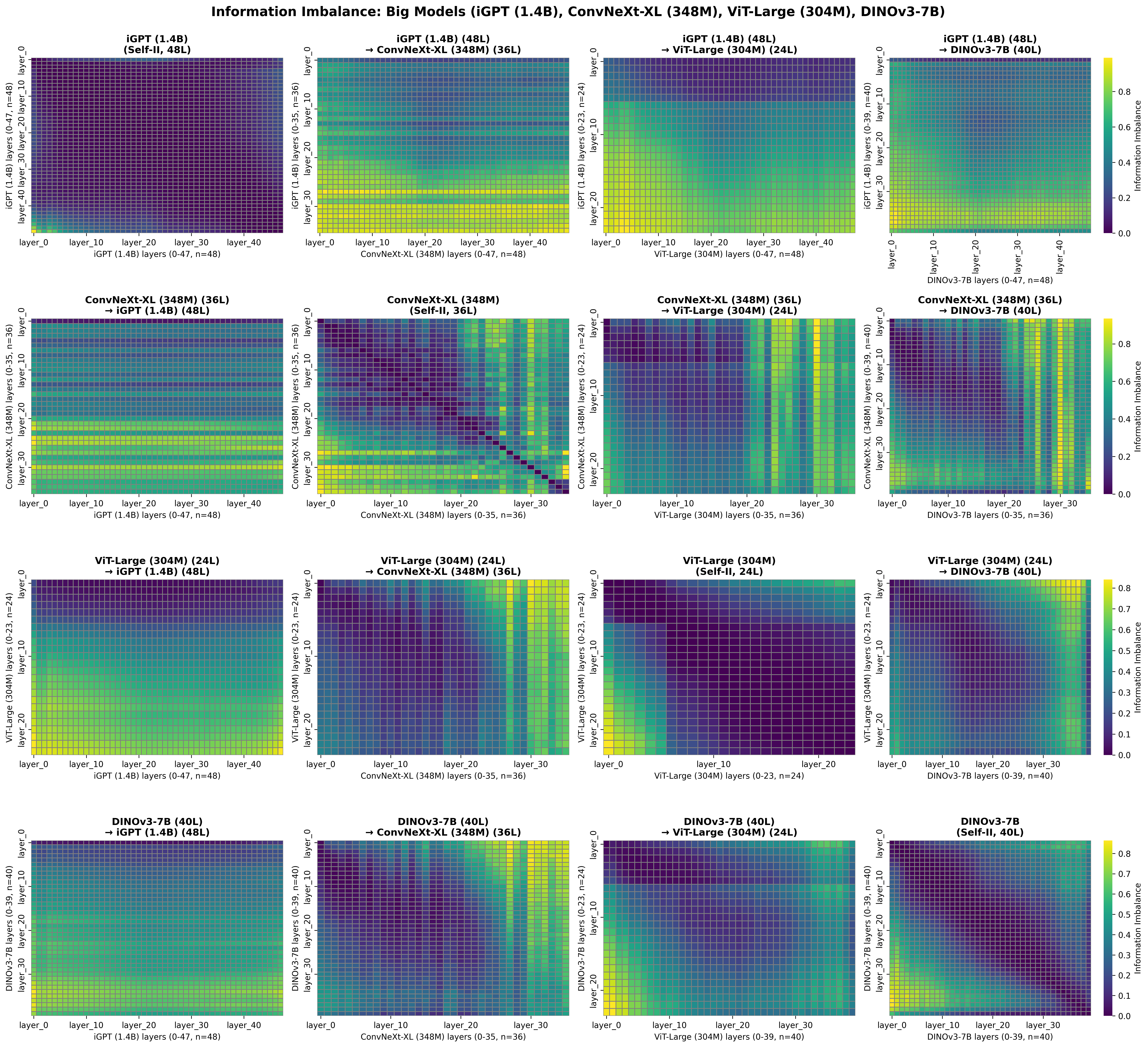}
    \caption{Information imbalance between models for \textbf{all layers}; lower information imbalance means that more neighbors are shared across image representations on the corresponding layers. Four models are compared, in both directions.  The title of each subgraph indicates the direction of information imbalance, with an arrow from the predicting model towards the predicted one.}
    \label{fig:big-grid}
\end{figure}

In Fig.~\ref{fig:big-grid}, instead of plotting an early, middle, and late layer for each of the bigger models, we repeat the analysis showing all layer-to-layer comparisons of all models. This figure is more information-heavy, but confirms previous results. Darker blue zones correspond to low information imbalance. Notably, iGPT has low information imbalance between nearly all its layers. When looking at cross-model comparisons, in all subplots apart from the diagonal, blue zones would be candidate \textit{universal} representations.
\begin{table*}[t]
\centering
\caption{Overview of Vision Models analyzed in this study. Each model varies in architecture, parameter count, and training objective. S, B, L, X denote Small, Base, Large, and Extra-Large variants respectively. Representative models (bold), with large parameter counts (underlined), are the ones used in the main text.}
\label{tab:model_summary_app}
\renewcommand{\arraystretch}{1.05}
\begin{tabular}{l l l l}
\toprule
\textbf{Model} & \textbf{Parameter Count (in millions)} & \textbf{Architecture} & \textbf{Training Objective} \\
\midrule
\textbf{iGPT} &  \underline{1362.1} & Vision Transformer & Next-pixel prediction \\
\textbf{DINOv2} & \underline{1136.4} & Vision Transformer & Self-supervised distillation \\
DINOv3 & 22.1 (S) / 86.6 (B) / 304.4 (L) / 1136.5 (XL) & Vision Transformer & Self-supervised distillation \\
DINOv3-ConvNeXt & 28.6 (S) / 88.6 (B) / 197.8 (L) & Convolutional N.N. & Self-supervised distillation \\
\textbf{ConvNeXt} & 28.6 (S) / 88.6 (B) / 197.8 (L) / \underline{350.2 (XL)} & Convolutional N.N. & Classification (ImageNet) \\
ConvNeXt V2 & 350.2 & Convolutional N.N. & FCMAE \\
\textbf{ViT} & 5.7 (S) / 86.6 (B) / \underline{304.3 (L)} & Vision Transformer & Classification (ImageNet) \\
\bottomrule
\end{tabular}
\end{table*}

\section{Sample-size, smoothness and repetition for information imbalance}
\label{app:std-ii}
\begin{figure}
    \centering
    \includegraphics[width=0.5\linewidth]{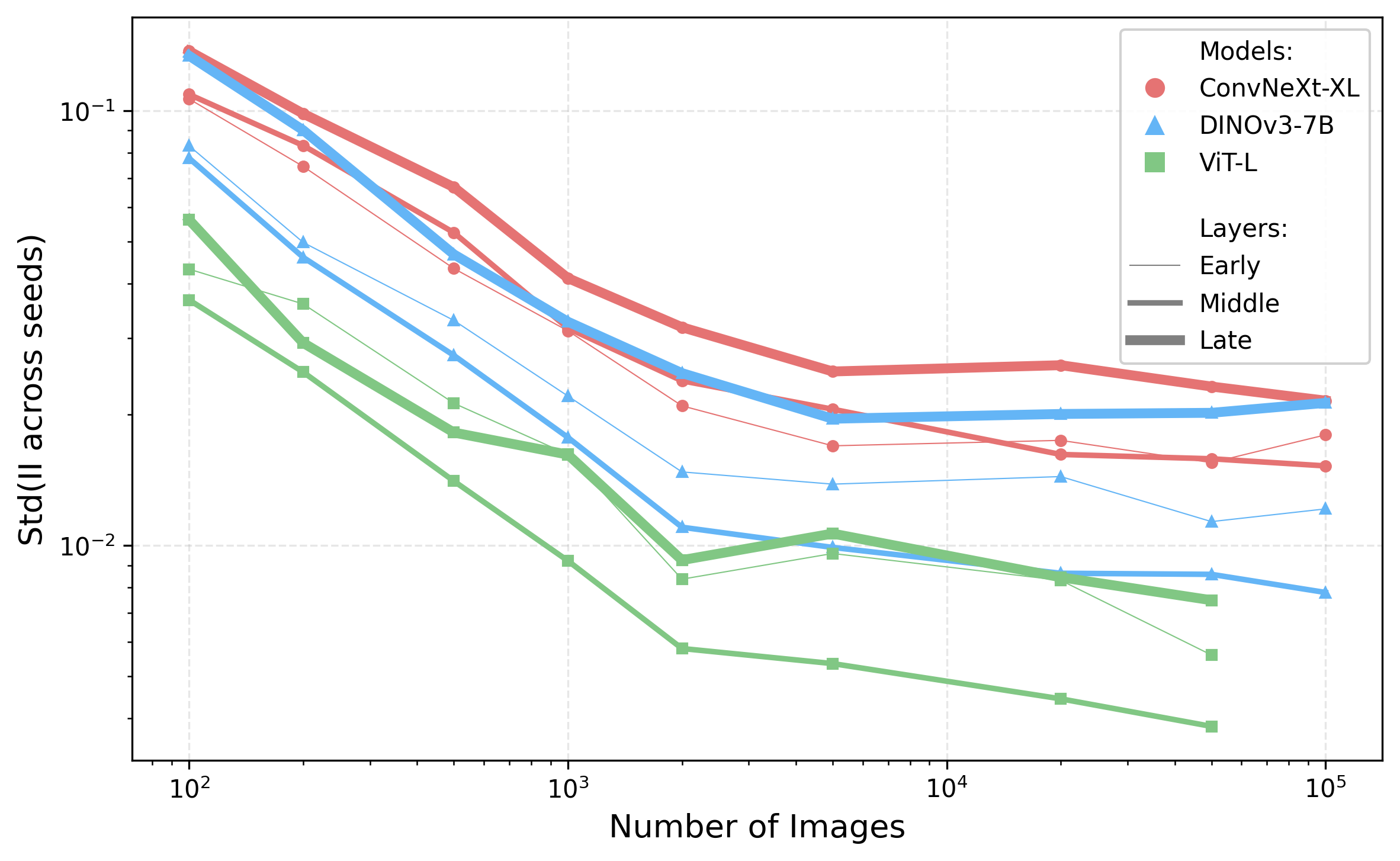}
    \caption{Standard deviation of information imbalance values across 10 increasingly large subsamples of the whole dataset , averaged over early, middle and late layers.}
    \label{fig:seed-ii}
\end{figure}

We now study the impact of the number of datapoints ($N$ in eq.~\ref{eq:ii}) on information imbalance. In Fig.~\ref{fig:seed-ii}, we compute information imbalance between layers from the same model on 10 different samples of the dataset, and report standard deviation between them. We seek to minimize the uncertainty due to sampling by choosing a big enough $N$. While standard deviation sharply decreases as $N$ grows at first, it becomes constant after $10^4$. This tells us that the remaining variation is not due to sample size. Among all models, ConvNeXt is the one that shows most variation, which is coherent with it being the least smooth model. ViT has the lowest information imbalance, probably due to the floor effect observed when predicting it's own later layers in Fig~\ref{fig:big_ii} where more than half of its information imbalance are 0.

\begin{figure}
    \centering
    \includegraphics[width=0.5\linewidth]{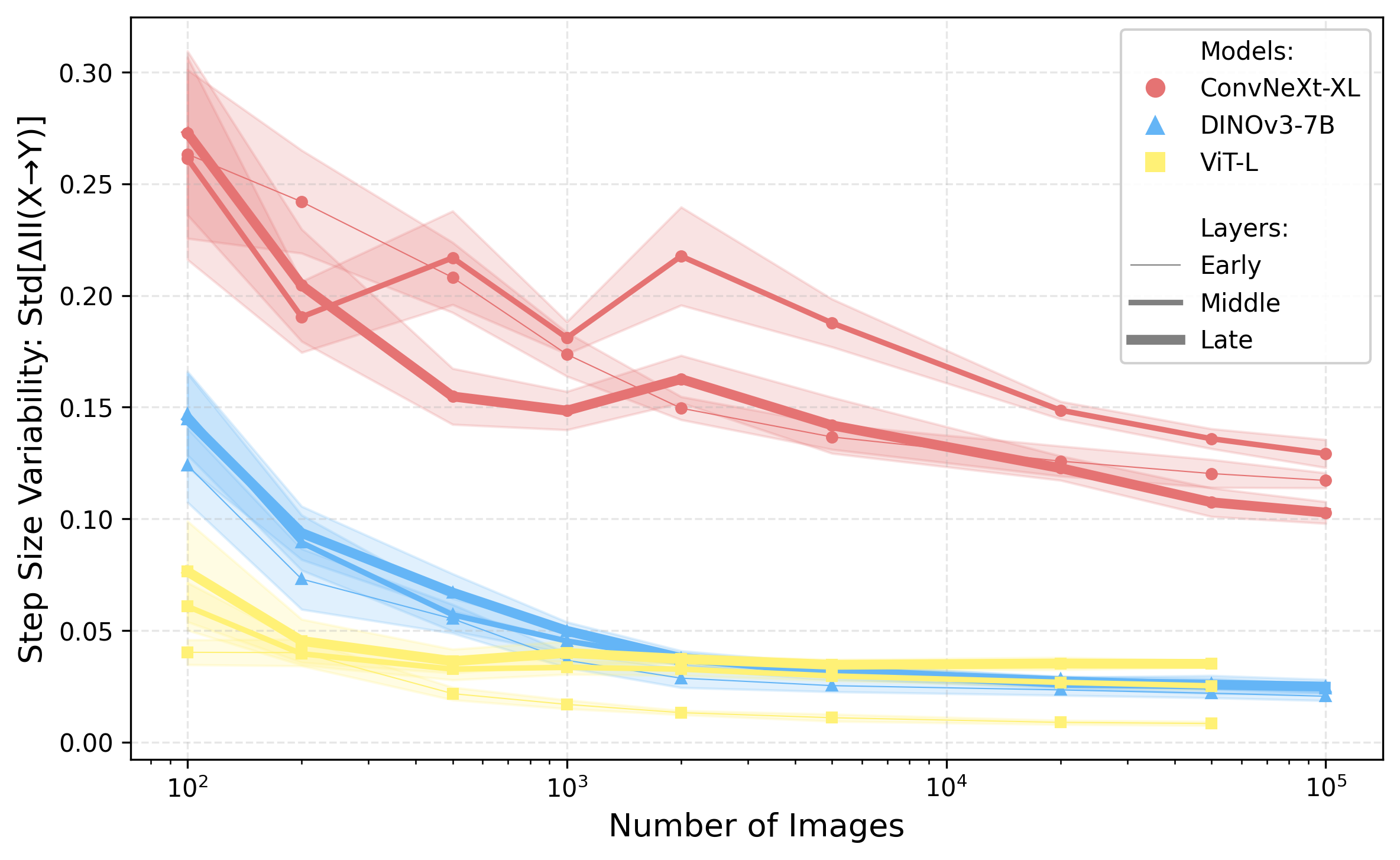}
    \caption{Standard deviation of the layerwise difference between information imbalance. Low values mean that information imbalance increases or decreases consistently, resulting in a very smooth curve. High values show irregularities, with one layer changing a lot and the next one very little. The shaded areas are standard deviations over ten seeds.}
    \label{fig:smoothness-ii}
\end{figure}

In Fig.~\ref{fig:smoothness-ii}, we test an additional hypothesis to explain the non-smooth information imbalance if ConvNeXt (Fig~\ref{fig:big_ii}).  We consider the possibility that the irregular steps in information imbalance between layers might be due to insufficient sample size, and that the curves would become smooth once sufficient sample size is provided. We test the effect of $N$ on smoothness. As in Sec.~\ref{sec:convn}, we define smoothness as the standard deviation of layer-to-layer differences. It emerges that ConvNeXt has the least smooth curve for all values of $N$. Additionally, the sample size effect is greatly reduced by $N=10^4$, and disappears by $N=10^5$. We therefore reject the hypothesis of sample-size playing a major role to explain non-smooth behavior of ConvNeXt.

\section{Out-Of-Domain Probing}
\label{app:probing}
\begin{figure}
    \centering
    \includegraphics[width=1\linewidth]{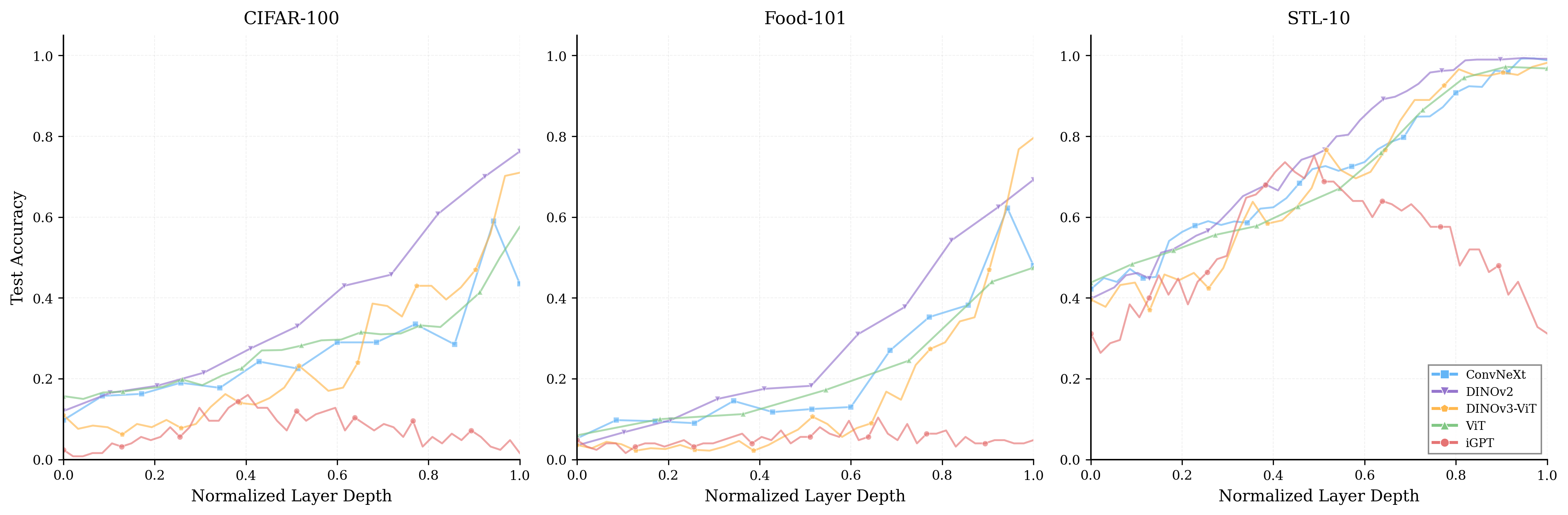}
    \caption{Classification accuracy on three out of domain datasets, using linear probes at each layer, for each model.}
    \label{fig:probing}
\end{figure}

We check the models' semantic ability on different out-of-domain tasks, at different depths. We train linear probes to classify images from three different out-of-domain datasets: CIFAR-100 \cite{krizhevsky2009learning}, made of low-resolution object images, Food-101 \cite{bossard2014food}, made of different food categories, and STL-10 \cite{coates2011analysis}, an unsupervised learning benchmark from which we only use the labeled test set.\footnote{Note that,  unlike in Sec.~\ref{sec:convn}, where we train a probe for each class, here we look at general classification accuracy. ConvNeXt results are therefore much smoother.}  Results are found in Fig.~\ref{fig:probing}. We replicate our finding that iGPT loses the semantic information that allows it to classify the images halfway through the model. The other models have similar incremental improvements of their probing accuracy as layer depth increases. %

\section{Low-level feature breakdown}
\label{app:low}
\begin{figure}
    \centering
    \includegraphics[width=1\linewidth]{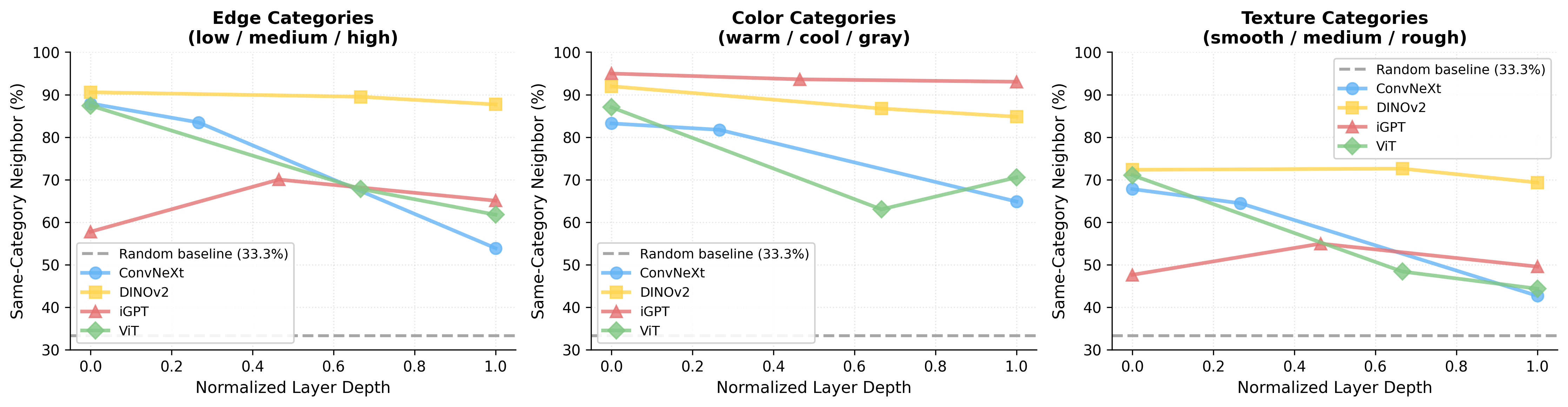}
    \caption{Percentage of images in an image 10 nearest neighbors that share one of the 3 discretized intensity ranges (high, medium, low) of a specific low-level property. Random baseline corresponds to the case where images are randomly spread in the space.}
    \label{fig:three_low_lev}
\end{figure}

We separate the low-level feature neighborhood analysis into the  three different properties in Fig.~\ref{fig:three_low_lev}. We are considering: amount of edges, color warmth, and texture categories. We find that the general trends observed in the main text hold in all categories. ViT and ConvNeXt show a progressive decrease in organisation according to low-level features, while iGPT and DINOv2 remain stable. iGPT even increases its proportion of same-category neighbors for the edge and texture features, barely overcoming ConvNeXt and ViT in the later layers, as it begins with much lower reliance on edge and texture features than all other models. Color reliance is high for all models, particularly for iGPT. DINOv2 is the most consistent across layers, staying within 5\% of its initial proportion of neighbors until the late layers.  

\section{Additional image neighborhood examples}
\label{app:neighbors}
\begin{figure*}[t]
    \centering
    \includegraphics[width=1\linewidth]{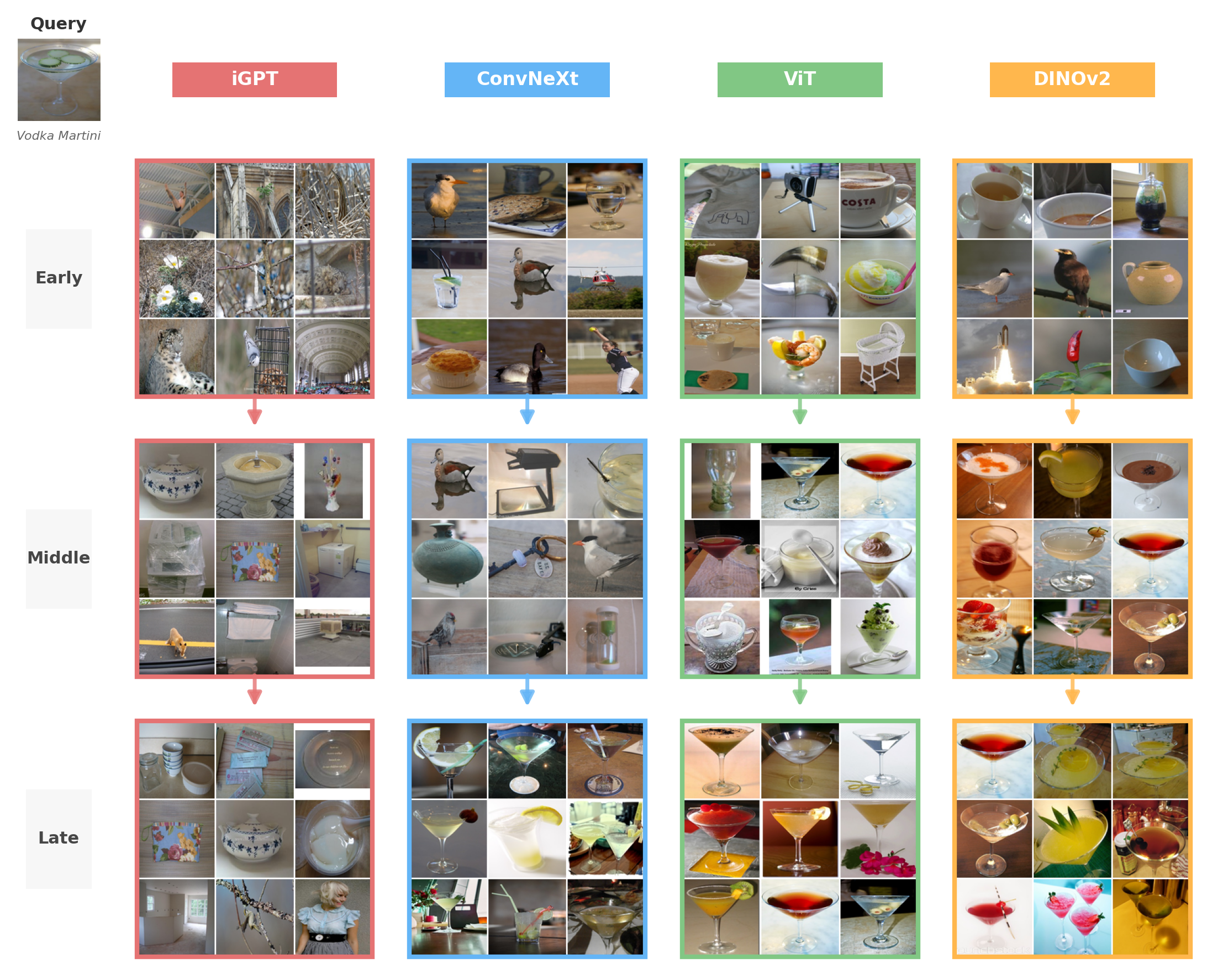}
    \caption{Nearest neighbors of a \textit{Vodka Martini} (top left) image from a set of 100k images, at different stages of different models. Each column is a model, each row is a specific layer. Early layers are all second layers, and late layers are all penultimate layers to avoid effects from tokenization or detokenization.}
    \label{fig:nn_cocktail}
\end{figure*}

In Fig.~\ref{fig:nn_cocktail}, there are different types of cocktail glasses for Convnext, while DINOv2 privileges images shot from the top. ViT seems most uniform. iGPT, as in the main-text example, does not extract the expected semantic information, though a clear change occurs between the early nature-like neighborhood and the later one with smooth backgrounds and a few container objects. 

\begin{figure*}[t]
    \centering
    \includegraphics[width=1\linewidth]{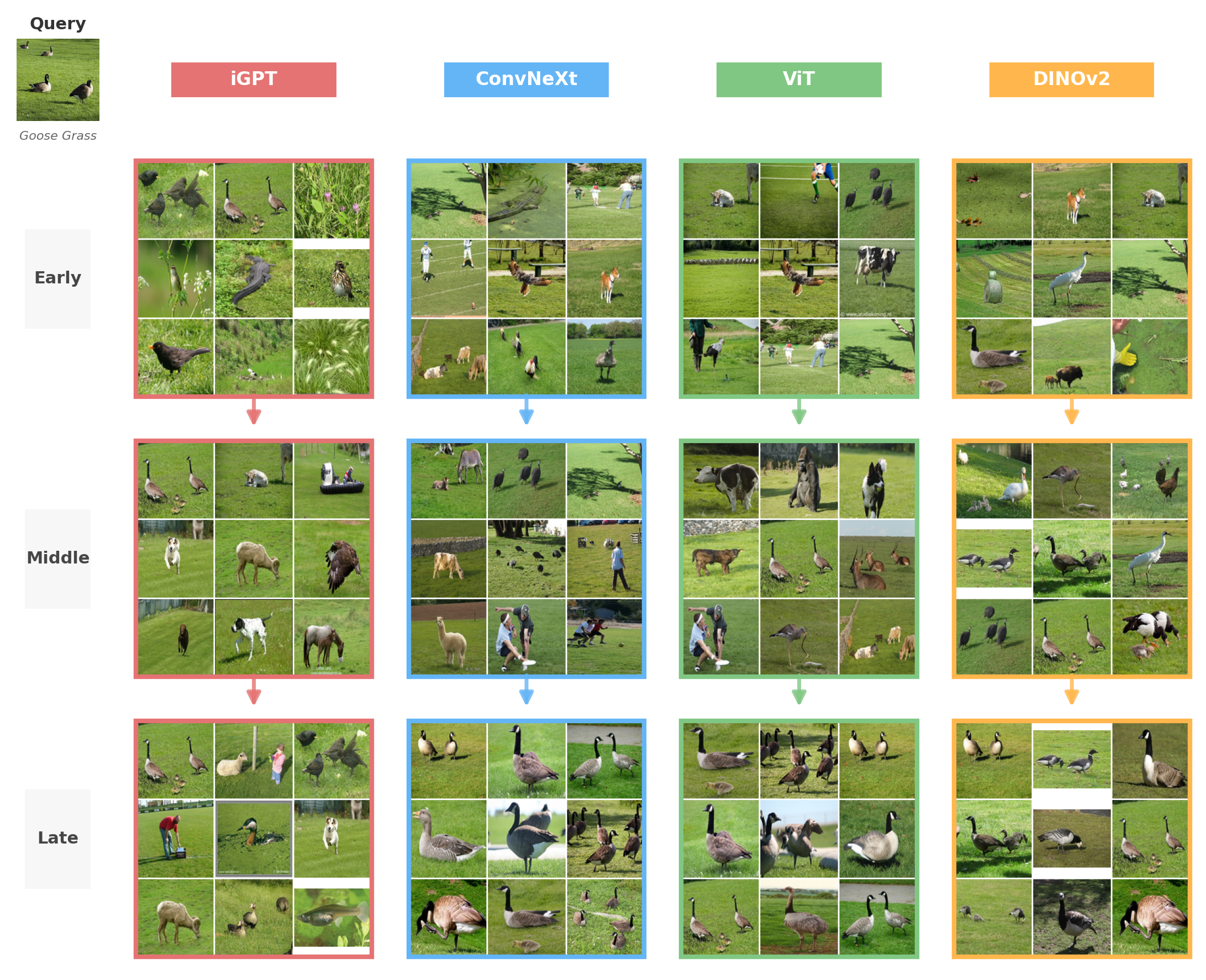}
    \caption{Nearest neighbors of a \textit{Goose on grass} (top left, imagenet class \textit{goose grass}) image from a set of 100k images, at different stages of different models. Each column is a model, each row is a specific layer. Early layers are all second layers, and late layers are all penultimate layers to avoid effects from tokenization or detokenization.}
    \label{fig:nn_ducks}
\end{figure*}

In the case of Fig.~\ref{fig:nn_ducks}, with a picture of geese on grass as query, early layers depict different types of lawns and animals, all with matching colors for the backgrounds. Except in the case of DINOv2, which has tightened the neighborhood to only birds (albeit of different kinds) by the middle layers, the other middle layers continue to consider other types of animals. It is worth noting that the nieghborhoods share images across models with an orange dog present for both ConvNeXt and DINOv2, or a landing eagle present for both ConvNeXt and ViT. Similarly, the two ultimate frisbee players are in the neighborhood of the middle layer for both ViT and ConvNeXt, and a flock of four black birds walking upwards are shared by ConvNext and DINOv2. In the final layers, the models all converge to nighborhoods made of the right species of bird (except for iGPT), and become slightly more tolerant to variations in the background.

\begin{figure*}[t]
    \centering
    \includegraphics[width=1\linewidth]{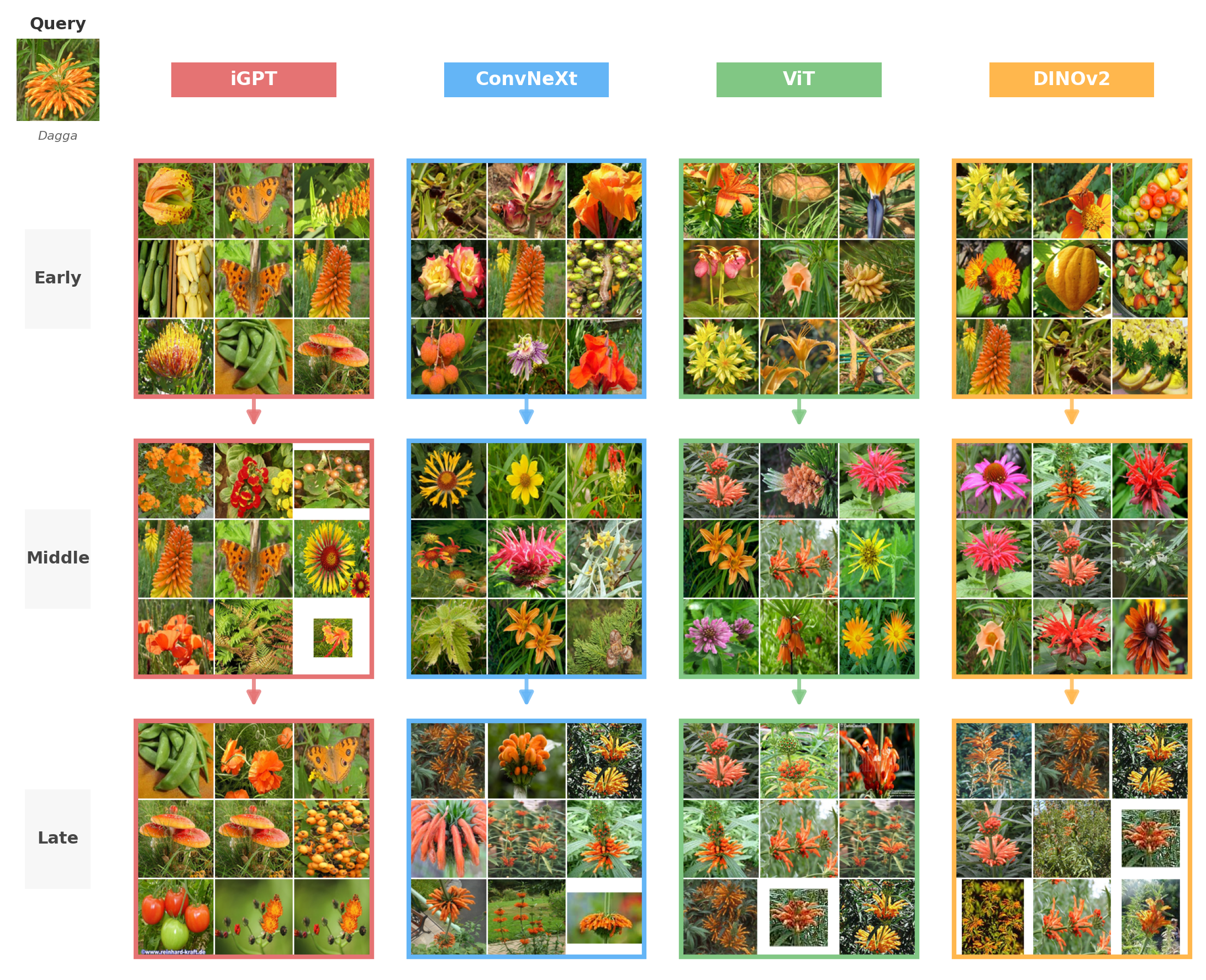}
    \caption{Nearest neighbors of a \textit{Dagga Flower} (top left) image from a set of 100k images, at different stages of different models. Each column is a model, each row is a specific layer. Early layers are all second layers, and late layers are all penultimate layers to avoid effects from tokenization or detokenization. Repeated images in the neighborhoods are duplicates in the Imagenet-21k dataset. Images with white borders have different scales and were also reduced in this format in the Imagenet21k dataset.}
    \label{fig:nn_dflower}
\end{figure*}

The Dagga Flower example in Fig.~\ref{fig:nn_dflower} works similarly to the previous examples. Early on, all images share colour and shape features. These might be fruit, butterflies, mushrooms. By the middle layers, models converge to only flowers, mostly photographs containing only one flower. They may have of different colors. While iGPT still has a butterfly, ViT and DINOv2 seem to have also identified a feature related to petal shape. By the final layer all models but IGPT have only flowers of the same species in the neighborhood, and a lot of the images are shared. It is worth noting that, for iGPT, the mushroom and beans were already part of the neighborhood in the early layer, and the butterfly was part of the neighborhood for all three layers.
\end{document}